\documentclass[11pt]{article}
\usepackage[utf8]{inputenc}
\usepackage[T1]{fontenc}
\usepackage[margin=1in]{geometry}
\usepackage{amsmath,amssymb,amsfonts}
\usepackage{booktabs,tabularx,array,longtable,multirow}
\usepackage{graphicx}
\usepackage{microtype}
\usepackage{enumitem}
\usepackage{xcolor}
\usepackage{hyperref}
\usepackage{tikz}
\usetikzlibrary{arrows.meta,positioning,calc}
\usepackage{seqsplit} 
\hypersetup{colorlinks=true,linkcolor=blue,citecolor=blue,urlcolor=blue}
\newcolumntype{Y}{>{\raggedright\arraybackslash}X}
\newcommand{\argmax}{\mathop{\mathrm{arg\,max}}}
\title{Proactive Knowledge Inquiry in Doctor-Patient Dialogue:\\Stateful Extraction, Belief Updating, and Path-Aware Action Planning}
\author{
Zhenhai Pan$^{1}$ \quad Yan Liu$^{1}$ \quad Jia You$^{1}$\\[0.6em]
$^{1}$The Hong Kong Polytechnic University, Hong Kong
}
\date{}
\begin{document}
\maketitle

\begin{abstract}
Most automated electronic medical record (EMR) pipelines remain output-oriented: they transcribe, extract, and summarize after the consultation, but they do not explicitly model what is already known, what is still missing, which uncertainty matters most, or what question or recommendation should come next. We formulate doctor-patient dialogue as a proactive knowledge-inquiry problem under partial observability. The proposed framework combines stateful extraction, sequential belief updating, gap-aware state modeling, hybrid retrieval over objectified medical knowledge, and a POMDP-lite action planner. Instead of treating the EMR as the only target artifact, the framework treats documentation as the structured projection of an ongoing inquiry loop. To make the formulation concrete, we report a controlled pilot evaluation on ten standardized multi-turn dialogues together with a 300-query retrieval benchmark aggregated across dialogues. On this pilot protocol, the full framework reaches 83.3\% coverage, 80.0\% risk recall, 81.4\% structural completeness, and lower redundancy than the chunk-only and template-heavy interactive baselines. These pilot results do not establish clinical generalization; rather, they suggest that proactive inquiry may be methodologically interesting under tightly controlled conditions and can be viewed as a conceptually appealing formulation worth further investigation for dialogue-based EMR generation. This work should be read as a pilot concept demonstration under a controlled simulated setting rather than as evidence of clinical deployment readiness. No implication of clinical deployment readiness, clinical safety, or real-world clinical utility should be inferred from this pilot protocol.
\end{abstract}

\section{Introduction}
Clinical consultations generate rich evidence: symptom descriptions, onset patterns, aggravating and relieving factors, prior history, test results, risk signals, and physician reasoning. Yet physicians must often conduct the consultation and document it at the same time. Recent progress in clinical note generation and dialogue summarization has substantially improved post hoc documentation support \cite{molenaar2020medical,abacha2023empirical,abacha2023mediqa}. However, most existing systems still follow a linear workflow: transcribe speech, extract information, and generate a final note after the fact. That workflow is useful for documentation assistance, but it does not explicitly reason about what is known so far, what remains missing, or what action would best close the most important gap while the consultation is still unfolding.

A clinician does not wait until the end of the encounter to decide what matters. The clinician maintains a running case model, identifies unresolved uncertainties, weighs risk, and chooses the next question or recommendation accordingly. We argue that a useful AI assistant should support exactly this process. In that sense, multi-turn doctor-patient dialogue is more naturally framed as a proactive knowledge-inquiry process than as a passive note-generation pipeline.

This paper proposes a unified framework for proactive knowledge inquiry in doctor-patient dialogue. The framework represents dialogue evidence as stateful events, maintains a structured \emph{CurrentState}, compares it against a target-oriented \emph{GoalState}, derives typed \emph{GapSignals}, retrieves supportive objects and reasoning paths from an objectified knowledge base, and selects the next action using a tractable POMDP-lite controller. Within this formulation, the EMR is not the control loop itself; it is a structured output view of the evolving internal state.

Our contributions are fourfold. First, we formulate dialogue-based EMR generation as a proactive inquiry problem under partial observability rather than a purely linear speech-to-record task. Second, we define a stateful extraction representation that distinguishes observed results, historical findings, recommendations, pending verification, negations, and related clinically meaningful evidence states. Third, we specify a unified inference-and-control framework that combines sequential belief updating, typed gap modeling, hybrid retrieval over objectified knowledge, and path-aware action planning. Fourth, we provide a controlled pilot protocol with raw-count denominators and stable audit units, together with a companion systems manuscript that instantiates the framework as an online prototype \cite{pan2026system}.

\begin{figure}[t]
    \centering
    \begin{tikzpicture}[font=\small, node distance=0.35cm and 0.35cm, >=Latex]
        \tikzstyle{box}=[draw, rounded corners, minimum height=0.9cm, align=center, inner sep=4pt]
        \tikzstyle{abox}=[draw, rounded corners, fill=blue!5, minimum height=0.9cm, align=center, inner sep=4pt]

        \node[box, minimum width=2.0cm] (s1) {Speech};
        \node[box, right=of s1, minimum width=2.0cm] (s2) {Text};
        \node[box, right=of s2, minimum width=2.1cm] (s3) {Fields};
        \node[box, right=of s3, minimum width=1.8cm] (s4) {EMR};
        \draw[->] (s1) -- (s2);
        \draw[->] (s2) -- (s3);
        \draw[->] (s3) -- (s4);
        \node[above=0.15cm of s2, font=\bfseries] {Linear pipeline};

        \node[abox, below=1.35cm of s1, minimum width=2.0cm] (p1) {Dialogue};
        \node[abox, right=of p1, minimum width=2.3cm] (p2) {Stateful\\Extraction};
        \node[abox, right=of p2, minimum width=2.55cm] (p3) {Current/Goal\\State + Gaps};
        \node[abox, right=of p3, minimum width=2.1cm] (p4) {Hybrid\\Retrieval};
        \node[abox, right=of p4, minimum width=2.2cm] (p5) {Action\\Planning};
        \node[abox, below=0.65cm of p3, minimum width=1.8cm] (p7) {EMR};
        \node[abox, below=1.55cm of p5, minimum width=2.25cm] (p6) {Updated\\Dialogue};
        \draw[->] (p1) -- (p2);
        \draw[->] (p2) -- (p3);
        \draw[->] (p3) -- (p4);
        \draw[->] (p4) -- (p5);
        \draw[->] (p5) -- (p6);
        \draw[->] (p3) -- (p7);
        \draw[->, dashed] (p6.west) -| (p1.south);
        \node[above=0.15cm of p3, font=\bfseries] {Proactive inquiry loop};
    \end{tikzpicture}
    \caption{Conventional linear EMR generation versus the proposed proactive inquiry loop. The main control loop updates dialogue state and feeds back into the ongoing interaction, while the EMR is generated as a structured output branch from the evolving state. The figure emphasizes that the proposed framework intervenes before note finalization by querying unresolved clinically relevant gaps.}
    \label{fig:linear-vs-proactive}
\end{figure}
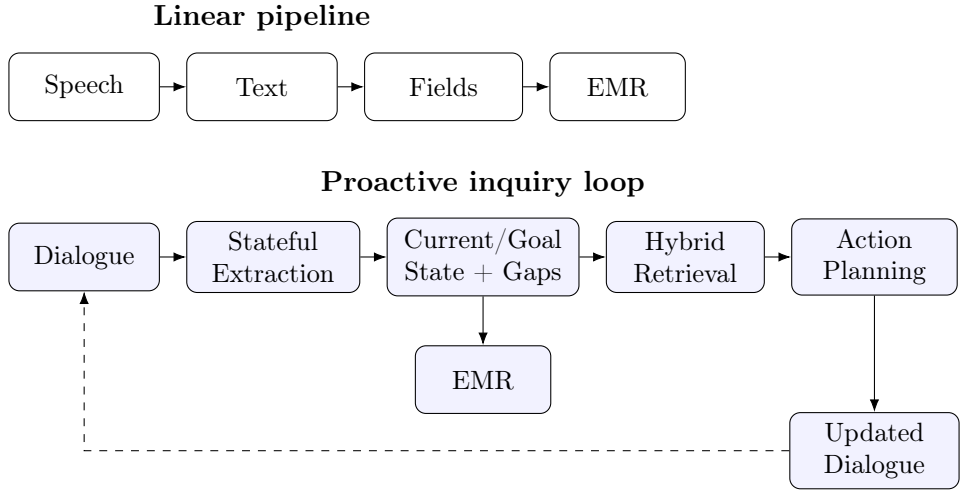

\section{Related Work}
\subsection{Clinical note generation and dialogue summarization}
Automated clinical documentation has evolved from dictation-oriented transcription toward dialogue summarization and note generation. Molenaar et al. proposed an early pipeline for medical dialogue summarization and automated reporting \cite{molenaar2020medical}. More recent work introduced larger open resources and empirical evaluations for clinical note generation from doctor-patient conversations \cite{abacha2023empirical}, while MEDIQA-Chat 2023 further consolidated the task definition and benchmark ecosystem \cite{abacha2023mediqa}. These studies are essential baselines for documentation quality, but they mostly optimize the final note rather than the interaction policy during the consultation.

\subsection{Clinical information extraction and dialogue state tracking}
Earlier reviews documented the emergence of clinical information extraction from EHR text \cite{meystre2008review}. More recent work has made strong progress in sequence labeling, concept normalization, and relation extraction over clinical free text \cite{jagannatha2016structured,frailenavarro2023clinicalie}. Dialogue state tracking (DST) provides a complementary line of work on maintaining structured state over multiple turns \cite{balaraman2021dst}. However, classical DST typically targets service-completion slots and fixed ontologies, whereas clinical dialogue requires stronger modeling of temporality, evidential status, risk-sensitive gaps, and action consequences.

\subsection{Retrieval-augmented assistance and healthcare knowledge graphs}
Retrieval-augmented generation provides a powerful general mechanism for grounding language models in external evidence \cite{lewis2020rag}. In medicine, knowledge graphs and structured clinical knowledge resources provide an additional layer of reasoning support beyond text chunks alone \cite{cui2023hkg}. Our framework moves from chunk-only retrieval toward a hybrid design that combines vector-based coarse retrieval, object-level reranking, and graph-path reasoning over clinically meaningful objects.

\subsection{Sequential decision making and uncertainty control}
The proposed controller is inspired by sequential decision making under uncertainty \cite{kaelbling1998pomdp}. We do not solve a full POMDP exactly; instead, we retain the practical elements most useful in clinical interaction: belief maintenance, typed gaps, and action scoring. Because these scores depend on model-derived uncertainty, calibration remains important \cite{guo2017calibration,angelopoulos2023conformal}.

\section{Problem Formulation}
\subsection{Dialogue as a partially observed inquiry process}
We consider a consultation as a multi-turn sequence of utterances
\begin{equation}
U_{1:T}=\{u_1,u_2,\ldots,u_T\},
\end{equation}
where each utterance may come from the patient, the physician, or another role such as a family member or an external report source. The underlying clinical state is only partially observable. At each turn, the system receives new observations and must update its internal representation of the case.

\subsection{Stateful events}
For an utterance $u_t$, the system extracts a set of stateful events
\begin{equation}
E_t=\{e_t^{(1)},e_t^{(2)},\ldots,e_t^{(n)}\},
\end{equation}
where each event is represented as
\begin{equation}
e_t^{(i)}=(f,v,s,\tau,r,\eta,c,\iota).
\end{equation}
Here, $f$ denotes the field name, $v$ the value, $s$ the state label, $\tau$ the temporal scope, $r$ the speaker role, $\eta$ the supporting evidence span, $c$ the confidence score, and $\iota$ a trace identifier. The central design choice is that the framework models the \emph{state} of each piece of evidence rather than only its semantic content.

\subsection{CurrentState, GoalState, and GapSignals}
At time $t$, the framework maintains a structured current state
\begin{equation}
C_t=\Phi(C_{t-1},E_t),
\end{equation}
and a dynamic goal state $G_t$ that represents the target information structure required by the evolving clinical scenario. The gap between current and target state is represented as
\begin{equation}
\Delta(C_t,G_t)=\{g_1,g_2,\ldots,g_m\},
\end{equation}
where each $g_i$ is a typed GapSignal such as an information gap, evidence gap, risk gap, differential gap, or path-blocking gap.

\subsection{Belief state}
Let $z_t$ denote the latent clinical hypothesis state at time $t$, and let $o_t$ denote the observation induced by utterance $u_t$. The system maintains a belief distribution
\begin{equation}
p(z_t\mid o_{1:t}),
\end{equation}
updated sequentially by
\begin{equation}
p(z_t\mid o_{1:t}) \propto p(o_t\mid z_t)\,p(z_t\mid o_{1:t-1}).
\end{equation}
In practice, the belief state is implemented as a structured object over candidate diagnoses, risks, and evidence sufficiency rather than as a fully exact probabilistic model.

\subsection{Action space and objective}
At each turn, the system generates a candidate action set
\begin{equation}
A_t=\{\texttt{ask},\texttt{verify},\texttt{explain},\texttt{recommend\_exam},\texttt{recommend\_plan},\ldots\},
\end{equation}
and selects the next action according to expected utility over uncertainty reduction, risk control, and path progress.

\section{Method}
\subsection{Stateful extraction}
Unlike conventional field extraction, which maps text to field-value pairs, our method requires the system to identify the state of each extracted item. We use a state taxonomy including {
\small\texttt{observed\_result},\allowbreak\enspace\texttt{confirmed},\allowbreak\enspace\texttt{verified},\allowbreak\enspace\texttt{completed},\allowbreak\enspace\texttt{historical\_result},\allowbreak\enspace\texttt{recommended},\allowbreak\enspace\texttt{\seqsplit{pending\_verification}},\allowbreak\enspace\texttt{unconfirmed},\allowbreak\enspace\texttt{not\_done},\allowbreak\enspace\texttt{negated},\allowbreak\enspace and\enspace\texttt{unknown}}. This distinction is critical because two utterances may mention the same concept while contributing very differently to downstream reasoning.

\begin{table}[t]
\centering
\caption{Representative examples from the state taxonomy used in stateful extraction.}
\label{tab:states}
\begin{tabularx}{\textwidth}{>{\raggedright\arraybackslash}p{0.22\textwidth}YY}
\toprule
State & Interpretation & Control implication \\
\midrule
observed\_result & Current observed finding or measured result & Strongly updates state and belief \\
historical\_result & Previously known finding or past exam result & Supports history with lower immediacy \\
recommended & Suggested future action & Triggers planning; not completed evidence \\
pending\_verification & Mentioned but unconfirmed & Gap source; often triggers verification \\
negated & Explicitly denied & Supports exclusion and contradiction handling \\
\bottomrule
\end{tabularx}
\end{table}

To reflect different control significance, each state is associated with a default state weight
\begin{equation}
w_s(e)=
\begin{cases}
1.0, & s\in\{\texttt{observed\_result},\texttt{confirmed},\texttt{verified}\},\\
0.7, & s=\texttt{completed},\\
0.5, & s=\texttt{historical\_result},\\
0.2, & s\in\{\texttt{recommended},\texttt{pending\_verification},\texttt{unconfirmed}\},\\
0.0, & s=\texttt{unknown}.\end{cases}
\end{equation}
These weights are engineering priors rather than universal clinical truths; they encode how strongly different evidence states should influence the current state and subsequent control decisions.

\subsection{Sequential belief updating}
Let $H=\{h_1,\ldots,h_n\}$ denote a set of candidate hypotheses. The uncertainty of the current belief state is quantified by entropy:
\begin{equation}
H(P)=-\sum_{i=1}^{n} p(h_i)\log p(h_i).
\end{equation}
To estimate the value of a candidate action $a$, we use expected information gain:
\begin{equation}
\mathrm{EIG}(a)=H(P_t)-\mathbb{E}_{o\sim p(o\mid a,P_t)}\left[H(P_{t+1})\right].
\end{equation}
This quantity measures how much uncertainty the system expects to reduce if action $a$ is taken.

\subsection{Hybrid retrieval over objectified knowledge}
Instead of indexing only text chunks, we parse medical documents, cases, guidelines, and rules into structured objects
\begin{equation}
O=\{o_1,o_2,\ldots,o_K\},
\end{equation}
including symptom units, diagnosis units, exam units, risk-rule units, and case-summary objects. Given the current state, the framework first performs vector-based coarse retrieval using a query embedding $q_t$ and candidate embeddings $e_j$:
\begin{equation}
\mathrm{sim}_{\cos}(q_t,e_j)=\frac{q_t\cdot e_j}{\|q_t\|\,\|e_j\|}.
\end{equation}
Coarse vector similarity alone is insufficient, so we rerank retrieved candidates at the object level using a composite score
\begin{equation}
S(C,o)=\alpha_1 S_f+\alpha_2 S_s+\alpha_3 S_g+\alpha_4 S_d+\alpha_5 S_r+\alpha_6 S_p+\alpha_7 S_{st},
\end{equation}
where $S_f$ denotes field-level similarity, $S_s$ structural similarity, $S_g$ graph-position similarity, $S_d$ distance to the goal state, $S_r$ risk alignment, $S_p$ path utility, and $S_{st}$ state compatibility. For a candidate reasoning path
\begin{equation}
\pi=(v_0,e_1,v_1,e_2,\ldots,e_L,v_L),
\end{equation}
we define path cost
\begin{equation}
\mathrm{Cost}(\pi)=\sum_{\ell=1}^{L} c(e_{\ell})+\rho\,\Omega(\pi),
\end{equation}
normalize it across candidate paths, and convert it into a higher-is-better path score. Vector-level, object-level, and path-level evidence are then fused into a final retrieval score.

\begin{figure}[t]
\centering
\begin{tikzpicture}[node distance=0.6cm and 0.5cm, every node/.style={font=\small}]
\tikzstyle{box}=[draw, rounded corners, minimum height=0.8cm, minimum width=2.2cm, align=center]
\node[box] (q) {Current state\\query};
\node[box, right=of q] (v) {Vector coarse\\retrieval};
\node[box, right=of v] (o) {Object-level\\reranking};
\node[box, right=of o] (p) {Path-aware\\reasoning};
\node[box, right=of p] (a) {Actionable\\support};
\draw[-{Latex[length=2mm]}] (q) -- (v);
\draw[-{Latex[length=2mm]}] (v) -- (o);
\draw[-{Latex[length=2mm]}] (o) -- (p);
\draw[-{Latex[length=2mm]}] (p) -- (a);
\end{tikzpicture}
\caption{Hybrid retrieval in the proposed framework: vector coarse retrieval narrows the candidate set, object-level reranking aligns candidates with state structure, and path-aware reasoning prioritizes clinically actionable support.}
\label{fig:retrieval}
\end{figure}
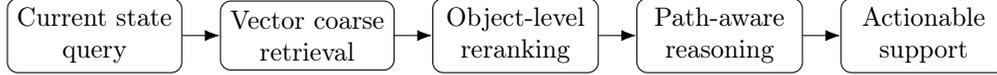

\subsection{POMDP-lite action planning}
A full POMDP is theoretically attractive but often impractical in this setting because of state-space complexity, uncertain transition models, and real-time constraints. We therefore retain the essential ideas of partial observability, belief maintenance, and action utility optimization while replacing exact global planning with a tractable local controller. Let the action utility be
\begin{equation}
U(a)=\lambda_1 IG(a)+\lambda_2 RR(a)+\lambda_3 PS(a)+\lambda_4 EG(a)-\lambda_5 RP(a)-\lambda_6 CL(a)+\lambda_7 CB(a),
\end{equation}
where $IG(a)$ is information gain, $RR(a)$ risk reduction, $PS(a)$ path shortening, $EG(a)$ explanation gain, $RP(a)$ redundancy penalty, $CL(a)$ cognitive-load penalty, and $CB(a)$ conservative bias. The selected action is
\begin{equation}
a_t^{\ast}=\argmax_{a\in A_t} U(a).
\end{equation}

\begin{table}[t]
\centering
\caption{Turn-level control loop of the proactive knowledge-inquiry controller.}
\label{tab:algo}
\begin{tabularx}{\textwidth}{>{\raggedright\arraybackslash}p{0.08\textwidth}Y}
\toprule
Step & Operation \\
\midrule
1 & Extract stateful events $E_t \leftarrow \mathrm{StatefulExtract}(u_t)$. \\
2 & Update current state $C_t \leftarrow \Phi(C_{t-1},E_t)$. \\
3 & Update belief $B_t \leftarrow \mathrm{BeliefUpdate}(B_{t-1},E_t)$. \\
4 & Derive gaps $\Delta_t \leftarrow \Delta(C_t,G_t)$. \\
5 & Retrieve supportive objects and candidate paths from knowledge base $K$. \\
6 & Generate candidate actions from state, belief, gaps, and retrieval outputs. \\
7 & Score each candidate action using information gain, risk reduction, path progression, explanation gain, redundancy penalty, cognitive-load penalty, and conservative bias. \\
8 & Select $a_t^{\ast}=\argmax_{a\in A_t} U(a)$ and return the updated state. \\
\bottomrule
\end{tabularx}
\end{table}

\section{Controlled Pilot Evaluation}
\subsection{Pilot setting}
We evaluate the framework on ten standardized multi-turn doctor-patient dialogues adapted from representative consultation scripts: four chest-discomfort cases, three upper-abdominal-pain cases, and three additional risk-sensitive scenarios. Each dialogue is associated with gold state-event annotations, gap annotations, target-state requirements, and risk items. Across the ten pilot dialogues, the audit includes 180 gold information items, 140 structure-critical record slots, and 60 critical-risk items. Retrieval is evaluated on 300 query points aggregated across dialogues rather than from a single conversation. The 300 retrieval queries are protocol-defined unresolved-state prompts derived from the gold schema of the ten pilot cases, rather than naturally occurring user queries from independent encounters.

\subsection{Baselines and metrics}
We compare four systems: (A) direct generation without explicit state tracking, (B) chunk-only retrieval-augmented generation without object-level or path-level reasoning, (C) rule-template questioning without belief-driven prioritization, and (D) the full proactive framework. We report five main indicators: Coverage, Risk Recall, Structural Completeness, Redundancy, and $T_{\mathrm{goal}}$.

Let $Y$ denote the set of gold target information items for a case and $\hat{Y}$ the subset correctly covered by the system. Let $R$ denote the set of gold critical-risk items and $\hat{R}$ the subset correctly surfaced by the system. We compute
\begin{equation}
\mathrm{Coverage}=\frac{|\hat{Y}|}{|Y|}, \qquad \mathrm{RiskRecall}=\frac{|\hat{R}|}{|R|}.
\end{equation}
For the final record, let $S$ denote the set of required structural slots and $S^{+}$ the subset filled with semantically correct values:
\begin{equation}
\mathrm{StructuralCompleteness}=\frac{|S^{+}|}{|S|}.
\end{equation}
If $A$ is the set of proposed actions and $A_{\mathrm{low}}$ the subset judged redundant or clinically low-value by the gold protocol, then
\begin{equation}
\mathrm{Redundancy}=\frac{|A_{\mathrm{low}}|}{|A|}.
\end{equation}
Finally, $T_{\mathrm{goal}}$ is the first turn at which all mandatory target-state slots and predefined risk checks have been satisfied.

\subsection{Overall method-level comparison}
Table~\ref{tab:overall} summarizes the pilot evaluation. Under the present protocol, the framework shows a suggestive directional pattern relative to passive generation, chunk-only retrieval, and template-heavy questioning. In raw counts, the full framework covers 150/180 gold information items, fills 114/140 structural slots correctly, surfaces 48/60 risk items, and produces 15 redundant prompts out of 95 total prompts.

\begin{table}[t]
\centering
\caption{Pilot evaluation under a controlled simulated setting. Higher is better for Coverage, Risk Recall, and Structural Completeness. Lower is better for Redundancy and $T_{\mathrm{goal}}$. Baseline A is non-interactive; therefore Redundancy and $T_{\mathrm{goal}}$ are not applicable.}
\label{tab:overall}
\begin{tabularx}{\textwidth}{Y>{\centering\arraybackslash}p{0.12\textwidth}>{\centering\arraybackslash}p{0.12\textwidth}>{\centering\arraybackslash}p{0.13\textwidth}>{\centering\arraybackslash}p{0.11\textwidth}>{\centering\arraybackslash}p{0.1\textwidth}}
\toprule
Method & Coverage (\%) & Risk Recall (\%) & Structural Completeness (\%) & Redundancy (\%) & $T_{\mathrm{goal}}$ \\
\midrule
Baseline A: Direct generation & 55.6 & 30.0 & 59.3 & N/A & N/A \\
Baseline B: Chunk-only RAG & 69.4 & 60.0 & 71.4 & 24.8 & 6.8 \\
Baseline C: Rule-template questioning & 77.8 & 70.0 & 78.6 & 28.4 & 7.4 \\
Ours & 83.3 & 80.0 & 81.4 & 15.8 & 5.8 \\
\bottomrule
\end{tabularx}
\end{table}

\subsection{Retrieval-layer comparison}
To isolate the contribution of the retrieval design, we compare chunk-level retrieval and the proposed hybrid retrieval on the 300-query benchmark. The hybrid design more reliably ranks clinically actionable evidence near the top. It corresponds to 261/300 successful top-5 recalls, 249/300 object hits, and 219/300 path hits, whereas chunk-only retrieval corresponds to 231/300, 214/300, and 162/300 respectively.

\begin{table}[t]
\centering
\caption{Retrieval results on a 300-query benchmark aggregated across dialogues.}
\label{tab:retrieval}
\begin{tabular}{lcc}
\toprule
Metric & Chunk-only RAG & Hybrid Retrieval \\
\midrule
Recall@5 & 0.770 & 0.870 \\
MRR@5 & 0.560 & 0.710 \\
nDCG@5 & 0.680 & 0.790 \\
Object hit rate & 0.71 & 0.83 \\
Path hit rate & 0.54 & 0.73 \\
\bottomrule
\end{tabular}
\end{table}

\subsection{Focused case study}
Consider a chest-discomfort scenario. Early in the dialogue, the patient reports chest tightness without sufficient discriminative detail. A chunk-only baseline can retrieve general chest-pain guidance but cannot robustly prioritize which evidence is most actionable. As the patient later adds exertion-related worsening and partial radiation symptoms, the proposed framework updates the belief state, detects unresolved risk gaps, reranks ECG-related and radiation-related objects higher, and prioritizes a risk-closing action rather than continuing broad generic questioning. This illustrates that the improvement does not come from any single module alone, but from the interaction among stateful evidence, belief dynamics, retrieval, and action selection.

\section{Relationship to the Companion Systems Paper}
This manuscript focuses on formulation: the state, gap, retrieval, and action abstractions; the mathematical structure of the controller; and the controlled pilot protocol that makes the problem auditable. Streaming ASR, punctuation restoration, belief stabilization under imperfect input, runtime budgeting, and online systems ablations are treated in the companion systems manuscript \cite{pan2026system}. The boundary is deliberate: the present paper argues for the \emph{problem formulation}, while the companion paper argues for the \emph{online system realization}.

\section{Discussion and Limitations}
The main contribution of this manuscript is methodological. By explicitly modeling evidence state, uncertainty, gaps, retrieval objects, and action utility, the framework turns automated documentation into an inquiry loop rather than a passive record-filling pipeline. On a small but fully specified benchmark, the proactive controller exhibits a suggestive trade-off pattern between thoroughness and efficiency relative to passive or template-heavy baselines within this pilot.

At the same time, the present evidence is a pilot study. The benchmark is small, scenario-based, and collected in a controlled simulated setting rather than in prospective clinic use. The protocol is intentionally structured and does not attempt to approximate the natural case distribution of outpatient encounters. The belief state is an approximation rather than a fully calibrated posterior, and future versions should strengthen calibration, abstention, and uncertainty-aware decision policies \cite{guo2017calibration,angelopoulos2023conformal}. The objectified knowledge layer also depends on parser quality; document routing, OCR, layout reconstruction, and table extraction can fail on complex clinical documents \cite{huang2022layoutlmv3,kim2021donut,smock2022pubtables}. We also do not report confidence intervals, multi-run variance, or annotation-agreement statistics in the present pilot. The present results should be interpreted as point estimates from a tightly controlled pilot protocol rather than as evidence of clinical generalizability. No implication of clinical deployment readiness, clinical safety, or real-world clinical utility should be inferred from this pilot. These limitations do not invalidate the formulation, but they do bound what can be claimed from the present pilot.

\section{Conclusion}
We presented a framework for proactive knowledge inquiry in doctor-patient dialogue, combining stateful extraction, sequential belief updating, hybrid retrieval over objectified knowledge, and POMDP-lite action planning. In this formulation, automated EMR generation is not treated as a terminal summarization step but as the structured output of an ongoing inquiry loop that reasons about what is known, what is missing, and what action should come next.

The controlled pilot study offers directionally supportive, but still suggestive, evidence that proactive inquiry may be a conceptually appealing formulation for dialogue-based EMR generation under a controlled protocol. It should be read as a pilot concept demonstration rather than as evidence of clinical readiness or generalizable clinical utility. The next stage is to test this formulation at larger scale under prospective clinical conditions and with stronger calibration, parser robustness, and multi-site evaluation.

\appendix
\section{Reference Annotation Schema}
Table~\ref{tab:schema} summarizes the gold schema used to define the audit units in the pilot protocol. Each dialogue turn can contribute one or more state events, and the final record is scored against normalized slot values rather than against surface strings alone.

\begin{longtable}{p{0.17\textwidth}p{0.28\textwidth}p{0.23\textwidth}p{0.22\textwidth}}
\caption{Reference gold schema used to define the audit units in the controlled pilot protocol.}\label{tab:schema}\\
\toprule
Field & Role in scoring & Example values & Used by metrics \\
\midrule
\endfirsthead
\toprule
Field & Role in scoring & Example values & Used by metrics \\
\midrule
\endhead
section & Record region / discourse region & HPI, ROS, Plan & Coverage, Structural Completeness \\
slot & Canonical clinical variable & cough, chest\_pain, chest\_xray & Coverage, Structural Completeness \\
normalized\_value & Normalized semantic value & present, absent, duration=3d & Coverage, state matching \\
status & Completion state & completed, recommended, not\_done & Coverage, action logic \\
temporality & Time anchoring & present, recent\_past, future & Coverage, risk interpretation \\
assertion & Polarity / certainty & positive, negative, proposed & Coverage, risk interpretation \\
risk\_flag & Whether the item is risk-critical & true / false & Risk Recall \\
evidence\_span & Linked utterance evidence & speaker, turn, character span & Audit traceability \\
\bottomrule
\end{longtable}

\begin{table}[h]
\centering
\caption{Illustrative protocol-defined retrieval query points from the pilot audit. These examples are derived from unresolved state gaps in the gold schema rather than from naturally occurring user queries.}
\label{tab:queryexamples}
\small
\begin{tabularx}{\textwidth}{p{0.20\textwidth}p{0.22\textwidth}Xp{0.14\textwidth}}
\toprule
Query point type & Source state / gap & Example prompt & Risk-critical? \\
\midrule
Allergy verification & Medication plan present but allergy status unresolved & ``Before recommending antibiotics, has penicillin allergy been explicitly confirmed?'' & Yes \\
Medication clarification & Drug mentioned without normalized dosage or frequency & ``What dosage and frequency should be recorded for the current antihypertensive medication?'' & No \\
Symptom duration follow-up & Chief complaint present but temporal detail missing & ``How long has the chest discomfort been present, and is it worsening?'' & Yes \\
Red-flag escalation check & Risk-sensitive symptoms partially observed but escalation status unresolved & ``Given persistent upper-abdominal pain with vomiting, should urgent testing or referral be prompted?'' & Yes \\
Exam / test completion status & Imaging or examination recommended but completion state unresolved & ``Was the chest X-ray already completed, or is it only being recommended at this stage?'' & No \\
\bottomrule
\end{tabularx}
\end{table}

\noindent In the pilot protocol, 60 of 180 audited information items (33.3\%) are risk-critical, while 120 of 180 (66.7\%) are non-risk-critical. Among the five illustrative query points in Table~\ref{tab:queryexamples}, three correspond to risk-critical gaps and two to non-risk-critical informational gaps. The query benchmark is therefore intentionally structured and partially enriched for clinically consequential unresolved states rather than intended to mirror the natural frequency distribution of outpatient encounters.

\end{document}